\pdfoutput=1

\documentclass[11pt]{article}

\usepackage[]{acl}
\usepackage{times}
\usepackage{latexsym}
\usepackage[T1]{fontenc}
\usepackage[utf8]{inputenc}
\usepackage{microtype}
\usepackage{multirow}
\usepackage{subcaption}
\usepackage{booktabs}
\usepackage{hyperref}
\usepackage{float}
\usepackage{covington}
\usepackage{adjustbox}
\usepackage{rotating}
\usepackage{comment}
\usepackage{paralist}

\title{The DURel Annotation Tool: \\Human and Computational Measurement of \\Semantic Proximity, Sense Clusters and Semantic Change}

\author{Dominik Schlechtweg$^{1}$, Shafqat Mumtaz Virk$^{2}$, Pauline Sander$^{1}$,\\\textbf{Emma Sköldberg$^{2}$, Lukas Theuer Linke$^{1}$, Tuo Zhang$^{1}$,}\\\textbf{Nina Tahmasebi$^{2}$, Jonas Kuhn$^{1}$, Sabine Schulte im Walde$^{1}$}\\$^{1}$University of Stuttgart, $^{2}$University of Gothenburg}

\begin{document}
\maketitle

\begin{abstract}
We present the DURel tool that implements the annotation of semantic proximity between uses of words into an online, open source interface. The tool supports standardized human annotation as well as computational annotation, building on recent advances with Word-in-Context models. Annotator judgments are clustered with automatic graph clustering techniques and visualized for analysis. This allows to measure word senses with simple and intuitive micro-task judgments between use pairs, requiring minimal preparation efforts. The tool offers additional functionalities to compare the agreement between annotators to guarantee the inter-subjectivity of the obtained judgments and to calculate summary statistics giving insights into sense frequency distributions, semantic variation or changes of senses over time.
\end{abstract}

\section{Introduction}
\label{sec:intro}

The concept of \textbf{semantic proximity} between word uses has a rather long tradition in Cognitive Semantics  \citep[][]{Blank97XVI} and is also acknowledged by notable lexicographers \citep[][]{Kilgarriff1997}. Semantic proximity quantifies how much the meanings of two word uses ``have in common'' \citep[][cf. p. 25]{Schlechtweg2023measurement}. The concept serves as (often vague) criterion in the lexicographic \textbf{clustering} process \citep[][]{Kilgarriff2006} and is thus essential to the process of creating dictionary entries of \textbf{word senses}. Being essential to the identification of word senses, semantic proximity has further relevance to research building on senses such as lexical semantic change or semantic variation \citep[][]{Schlechtweg2023measurement}.

In lexical semantics, multiple approaches operationalized semantic proximity in \textbf{human annotation} studies, showing that the concept can be practically implemented with reasonable agreement between annotators and correspondence to alternative annotation procedures \citep[][]{SoaresdaSilva1992,Erk13,Schlechtwegetal18}. 

Recently, there has been an upsurge on research in computational modeling of  semantic proximity between word uses under the name of \textbf{Word-in-Context} models \citep[][]{pilehvar2019wic,Armendariz19}, resulting from advances in modeling the meaning of word uses with contextualized embeddings \citep[][]{peters-etal-2018-deep,devlin-etal-2019-bert}. These models  reach high performance \citep{he2020deberta,raffel2020exploring} and thus serve as an excellent starting point for any practical task building on semantic proximity like creating dictionary entries, finding novel/non-recorded senses or identifying words that change their meaning.

We present the \textbf{DURel tool} combining the above-described lines of research into a user-friendly, online annotation interface with open source code.\footnote{\url{https://www.ims.uni-stuttgart.de/data/durel-tool}} The basic annotation data gathered in the system are judgments of semantic proximity between word uses \citep[][]{Blank97XVI,Erk13} from multiple human or computational annotators. The tool facilitates the annotation task by providing a data inspection interface and automatic data validation for researchers, an intuitive task interface for annotators, guidelines in multiple languages for annotator training as well as tutorial data for annotator testing. Computational annotators are provided by optimized Word-in-Context models \citep[][]{pilehvar2019wic,Arefyev2021Deep} trained on human semantic proximity judgments \citep[i.a.][]{Schlechtweg2021dwug,kutuzov2021threepart,Zamora2022lscd}. Semantic proximity judgments are represented in a graph \citep[][]{carthy16}, clustered with an automatic \textbf{graph clustering} technique \citep[][]{schlechtweg-etal-2020-semeval,Schlechtweg2021dwug} and visualized for analysis \citep[][]{thesis11}. This allows to measure word senses from \textbf{simple} and \textbf{intuitive} semantic proximity judgments between use pairs. The tool offers functionalities to compare the agreement between annotators to guarantee the \textbf{inter-subjectivity} of the obtained judgments. It provides further functionalities to calculate summary statistics over the annotated data giving insights into sense frequency distributions, semantic variation or changes of senses over time. The computational annotator component allows to generate word sense clusters for large sets of words and word uses, making it possible, for instance, to analyze large amounts of data or to search unlabelled data systematically for new senses.

\section{Related Work}
\label{sec:related}

We now compare DURel to existing text annotation tools and related tools from electronic lexicography. There is a number of general-purpose text annotation tools such as CATMA \citep{evelyn_gius_2022_6419805}, INCEpTION \citep{tubiblio106270}, MTURK\footnote{\url{https://www.mturk.com/}}, PhiTag\footnote{\url{https://www.ims.uni-stuttgart.de/data/phitag}}, POTATO \citep{pei2022potato} or Toloka\footnote{\url{https://toloka.ai/en/docs/}}.
Many of these allow to define a wide range of custom tasks and can in principle cover use pair annotation \citep[cf. e.g.][]{giulianelli-etal-2020-analysing,rushifteval2021}. However, this often requires preparation efforts including the writing of small programs as well as the formulation of guidelines and data for annotator training. The aim of the DURel tool is to minimize such additional efforts around organizing a use pair annotation study. DURel achieves this by focusing on this particular task, implementing standardized procedures which have proven to work well in previous studies \citep[i.a.][]{Schlechtwegetal18,haettySurel-2019,schlechtweg-etal-2020-semeval,Baldissin2022diawug}. Further unique features of the tool are the task-specific data analysis (see Section \ref{sec:visualization}) and computational annotators (see Section \ref{sec:pair}). Other tools, while offering annotation for a range of tasks, cannot offer these specific possibilities.

With its focus on a semantic annotation task (semantic proximity) and the sense inference functionality (see Section \ref{sec:stats}), the DURel tool is relevant to lexicography. The most widely used lexicographic tool is Sketch Engine.\footnote{\url{https://www.sketchengine.eu}} The DURel tool differs from Sketch Engine by focusing primarily on crowdsourcing human annotations and offering computational annotation models, whereas Sketch Engine offers frequency-based corpus analysis and manual dictionary making. The two tools could be integrated to provide improved analysis of word meaning.

\begin{table}[t]
\parbox{.45\linewidth}{
\centering
\tabcolsep=0.11cm
\begin{tabular}{ll}
\multirow{4}{*}{$\Bigg\uparrow$} &4: Identical\\
 &3: Closely Related\\
 &2: Distantly Related\\
 &1: Unrelated\\
\end{tabular}
\label{tab:scale2}}
\hfill
\parbox{.45\linewidth}{
\centering
\tabcolsep=0.11cm
\begin{tabular}{ll}
\multirow{4}{*}{$\Bigg\uparrow$}&Identity\\
&Context Variance\\
&Polysemy\\
&Homonymy
\end{tabular}
\label{tab:blank}}
\caption{The DURel relatedness scale \citep[][]{Schlechtwegetal18} on the left and its interpretation from \citet[][p. 33]{Schlechtweg2023measurement} on the right.}\label{tab:scale}
\end{table}

\begin{figure*}[t]
    \begin{subfigure}{0.33\textwidth}
\frame {        \includegraphics[width=\linewidth]{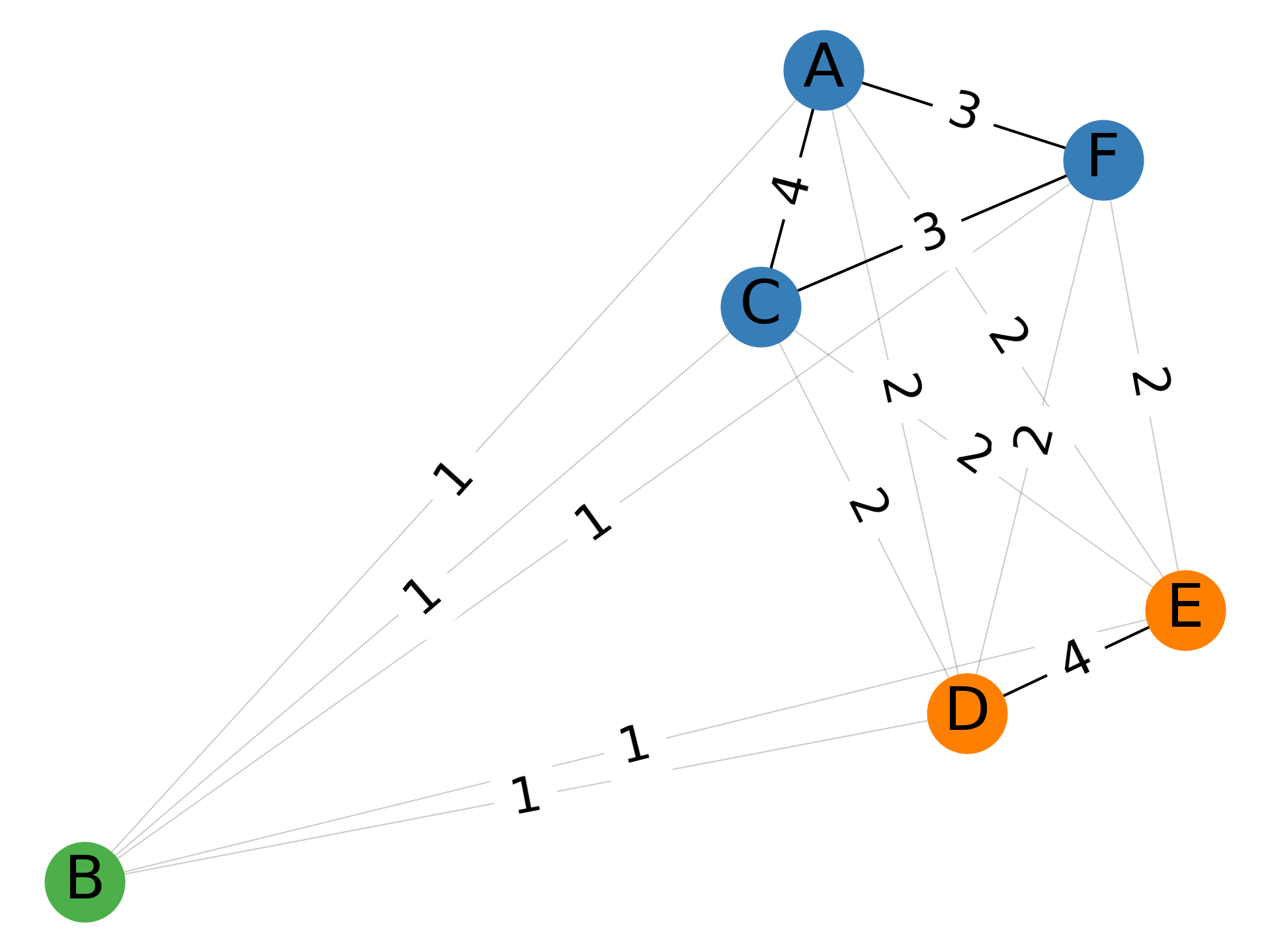}}
        \caption*{$G$}
      \end{subfigure}
    \begin{subfigure}{0.33\textwidth}
\frame{        \includegraphics[width=\linewidth]{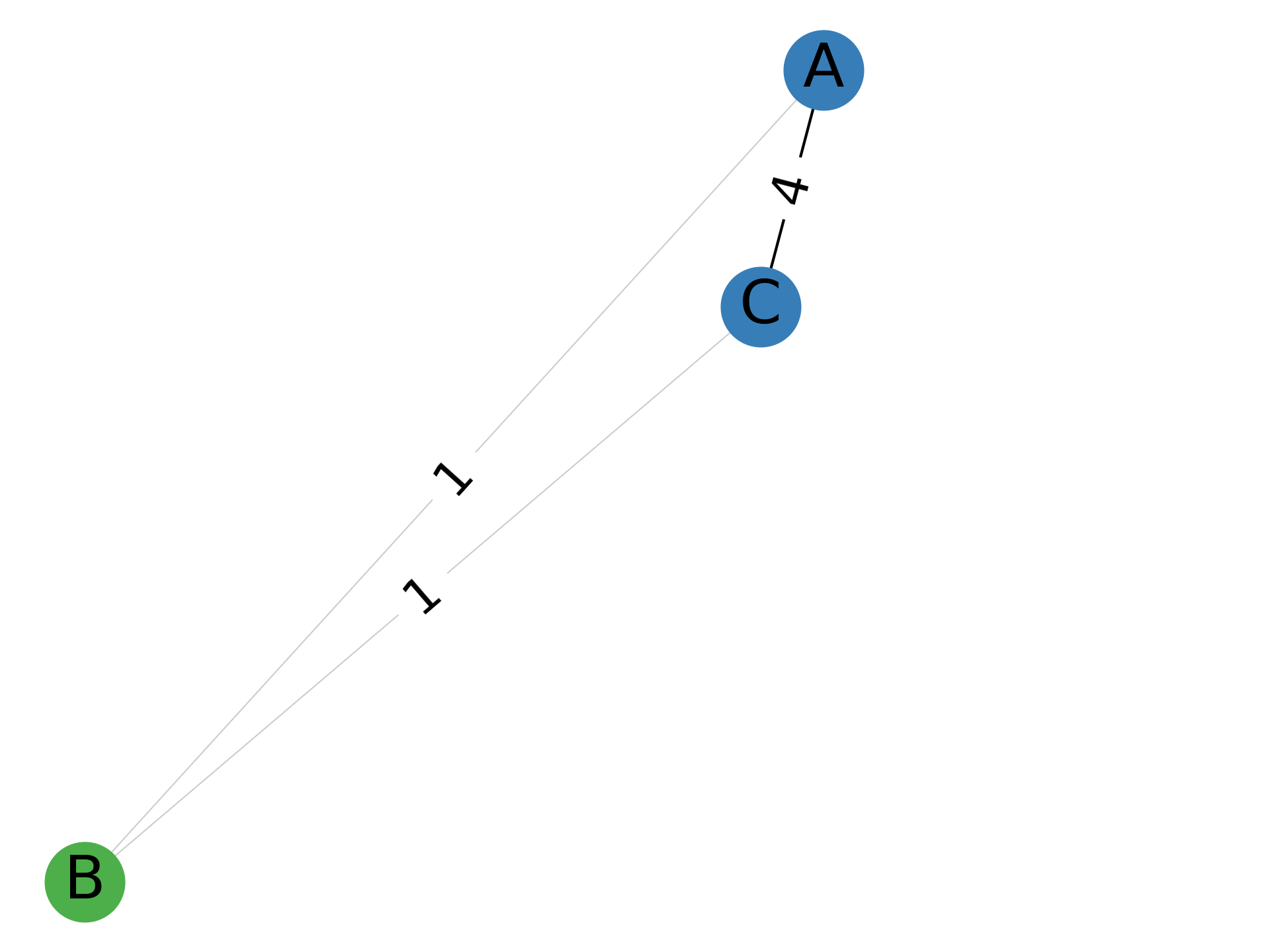}}
        \caption*{$G_1$}
    \end{subfigure}
    \begin{subfigure}{0.33\textwidth}
\frame{        \includegraphics[width=\linewidth]{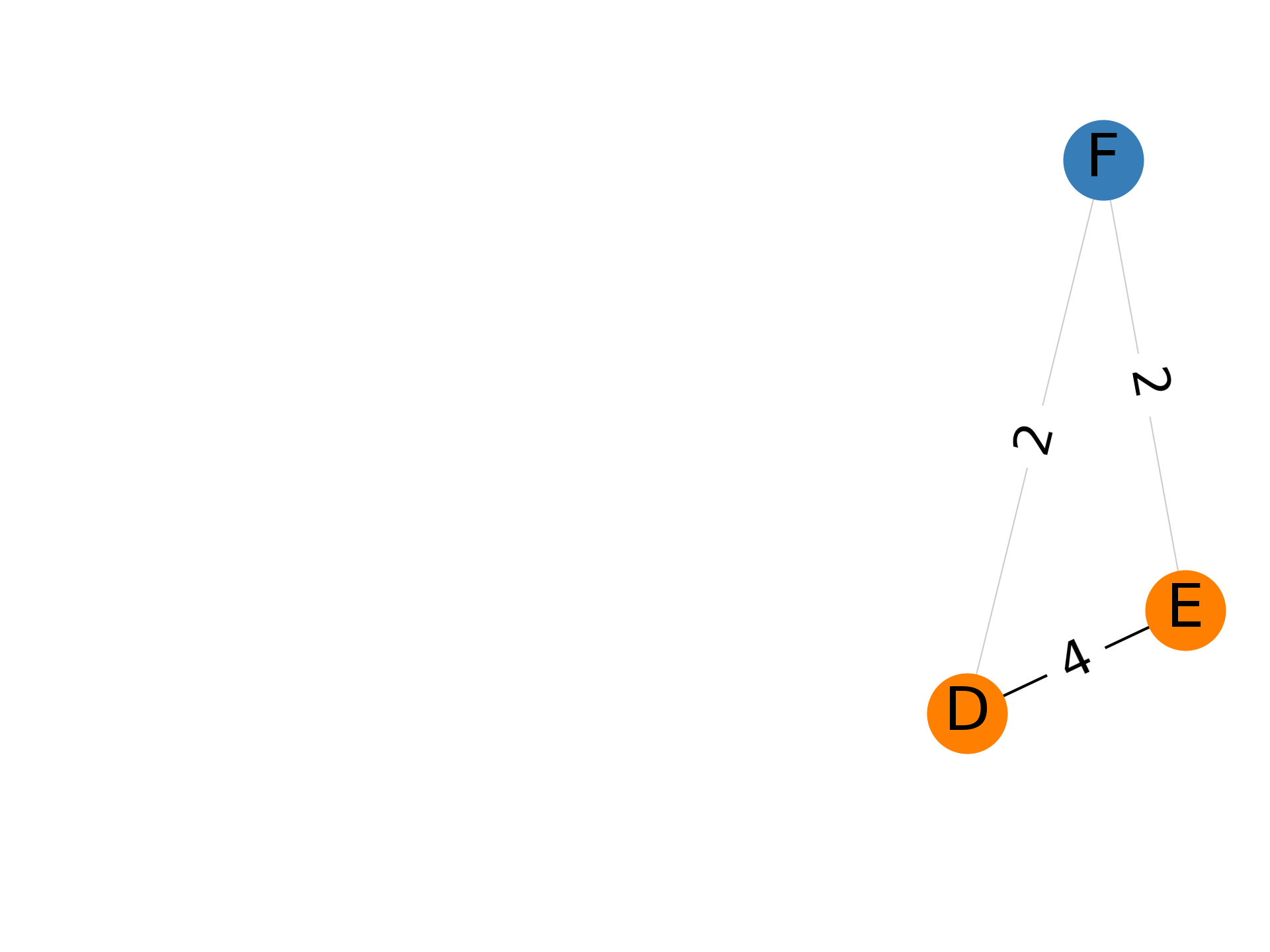}}
        \caption*{$G_2$}
    \end{subfigure}
   \caption{Clustered WUG $G$ of \textit{arm} (left), subgraph for 1st time period $G_1$ (middle) and subgraph for 2nd time period $G_2$ (right). \textbf{black}/\textcolor{gray}{gray} lines for \textbf{high ($\geq 2.5$)}/\textcolor{gray}{low ($< 2.5$)} edge weights. Spatial proximity of nodes loosely corresponds to their semantic proximity annotation. Visualization taken from \citet[][p. 40]{Schlechtweg2023measurement}.}\label{fig:graphsample1}
\end{figure*}

\section{Background}
\label{sec:background}
The DURel tool implements the word sense annotation scheme developed in \citet[][]{Schlechtwegetal18,schlechtweg-etal-2020-semeval,Schlechtweg2021dwug} and described in detail in \citet[][pp. 31ff.]{Schlechtweg2023measurement}. The scheme builds on use pair proximity annotation on a relatedness scale combined with a graph clustering procedure. Annotators are asked to judge the semantic relatedness of use pairs, such as the two uses of \textit{arm} in (\ref{ex:app1}) and (\ref{ex:app2}), on the scale in Table \ref{tab:scale}. 
\begin{example}\label{ex:app1}
[\ldots] she opened a vein in her little \textbf{arm}, and dipping a feather in the blood [\ldots] 
\end{example}%
\begin{example}\label{ex:app2}
[\ldots] he saw her within reach of his \textbf{arm}, yet the light of her eyes seemed as far off [\ldots]
\end{example}
The annotated data of a word is then represented in a graph, which we call Word Usage Graph (WUG), where vertices represent word uses, and weights on edges represent the (median) semantic relatedness judgment of a use pair. The final WUGs are clustered with Correlation Clustering \citep[][]{Bansal04,schlechtweg-etal-2020-semeval}.

Consider the example in Figure \ref{fig:graphsample1} to understand how senses and semantic change can be inferred with the DURel annotation procedure: Assume that WUG $G$ represents the semantic proximity structure annotated for the set of word uses $U$ of the English word \textit{arm} displayed in Table \ref{tab:corpus}.
The uses $U_1 = \{A,B,C\}$ and $U_2 = \{D,E,F\}$ were sampled from the two time periods 1820--1860 and 1950--1990 respectively ($t_1$, $t_2$). We derive sense clusters on $G$ by building three clusters of uses with high semantic proximity and low semantic proximity to other clusters: $C_1 = \{A,C,F\}$ (blue), $C_2 = \{D,E\}$ (orange), $C_3 = \{B\}$ (green). We then build the time-specific subgraphs $G_1$ and $G_2$ and are now able to compare the clusters between time periods. For instance, $C_3$ only exists in the first time period while $C_2$ only exists in the second time period.

\section{Tool description}

DURel is a web application supporting user interaction by browser (Mozilla Firefox is recommended).\footnote{Find a demo video at \url{https://www2.ims.uni-stuttgart.de/video/durel-tool/231121-durel-tool-demo.mp4}.} It was created on an architecture incorporating
Java Spring for backend, PostgreSQL for the database, HTML/CSS/JavaScript plus Thymeleaf for frontend, and a CSV format for transferring annotations. The source code of the tool is publicly available under a Creative Commons Attribution-NonCommercial-ShareAlike 4.0 International License.\footnote{\url{https://github.com/ChangeIsKey/durel_tool}}

\subsection{Data format}
Sets of a word uses as in Table \ref{tab:corpus} are the basic data type that DURel works with. These can be sampled from any corpus by the user and then uploaded to the tool (after registration) as one CSV per word through an interface. The files should contain at least one use context per line along with the target word and target sentence character indices. Additionally, they can contain meta-information such as the target word's POS, the date of the use or a user-specified grouping tag for uses, which can be used later for statistics and visualization. 

\begin{table*}[t]
\centering
\begin{adjustbox}{width=0.9\textwidth}
\frame{
 \begin{tabular}{l l p{0.9\textwidth}}
  \label{use:A} & 1824 & and taking a knife from her pocket, she opened a vein in her little \textbf{arm},
   \\ 
   \label{use:B} & 1842 & And those who remained at home had been heavily taxed to pay for the \textbf{arms}, ammunition;
   \\ 
   \label{use:C} & 1860 &   
   and though he saw her within reach of his \textbf{arm}, yet the light of her eyes seemed as far off \par
   \ldots  \\ 
  \label{use:D} & 1953 &  
  overlooking an \textbf{arm} of the sea which, at low tide, was a black and stinking mud-flat \\
   \label{use:E} & 1975 & twelve miles of coastline lies in the southwest on the Gulf of Aqaba, an \textbf{arm} of the Red Sea. 
   \\
  \label{use:F} & 1985 &   
  when the disembodied \textbf{arm} of the Statue of Liberty jets spectacularly out of the
\end{tabular}
}
\end{adjustbox}
\caption{Sample of diachronic corpus taken from \citet[][p. 41]{Schlechtweg2023measurement}.}\label{tab:corpus}
\end{table*}

\subsection{Features}
We now describe the central features of the DURel tool.

\subsubsection{Project management}
By clicking on the tab "Upload Project" the user can create an annotation project. He has to specify the project language and choose a set of CSV files containing word uses to upload. He has then two options options to create annotation instances (use pairs): (i) Let the system generate random sequences per annotator of all possible combinations of use pairs per word. (ii) Upload a user-defined sequence of annotation pairs, which will be presented to each annotator in randomized order. In both cases, the sequence within use pairs is swapped with a 0.5 probability. Alternatively, the user can upload his own gold annotations as annotation project.

Under the tab "My Projects", users can manage, download or delete their projects, as well as assign specific users to projects or make a project entirely public (see Figure \ref{fig:projects}). A user that has been granted access can annotate for that project. 

\subsubsection{Use pair annotation}
\label{sec:pair}
Use pairs are judged on the semantic proximity scale shown in Table \ref{tab:scale}. An annotator has the option of assigning a label between 1 (unrelated) and 4 (identical), or to assign no label (Cannot decide). 

\paragraph{Human annotation}
When an annotator registers to the DURel tool, they first need to successfully complete an annotation tutorial. Before starting the tutorial, the annotator is asked to read the guidelines page, which explains how to make semantic proximity judgments. In the tutorial, the annotator annotates only a few use pairs of different words. These judgments are then checked against a hidden gold standard. The annotator only passes the tutorial if he reaches a certain level of agreement. 

After passing the tutorial, annotators can annotate for  projects to which they have been assigned by clicking on the "Annotate" tab (see Figure \ref{fig:annotate} in Appendix).
Each project is divided into several words. Annotators can decide which word they annotate and the annotation can be paused at any time.

For each annotation instance, the tool records the judgment label, an optional comment, the annotator name and the timestamp of the judgment.

\paragraph{Computational annotation}
We use optimized, multi-lingual Word-in-Context (WiC) models as computational annotators. These are treated analogous to human annotators in the DURel system. They appear as users/annotators in the project management and can be assigned to an annotation project or an individual word by the user through clicking on the "Tasks" tab. Creating a task with a computational annotator will trigger an automatic annotation pipeline in the DURel backend retrieving annotation labels with the respective model and storing them in the DURel database. This allows large-scale data labeling for uploaded projects. The currently available computational annotators are:
\begin{itemize}
   \setlength\itemsep{0pt}
    \item \textbf{Random} samples a random integer between 1 and 4 with uniform probability (as baseline).
    \item \textbf{XLMR+MLP+Binary}: XLMR \citep{conneau-etal-2020-unsupervised} vectorizer with multi-layer perception and binary classification head on concatenated vectors; trained on WiC dataset; predicts either value 1 or 4. 
    \item \textbf{XL-Lexeme}: bi-encoder that vectorizes the input sequences using a XLMR-based Siamese Network \citep{Cassotti2023Xllexeme}; trained to minimize the contrastive loss with cosine distance on several WiC datasets; predicts either value 1 or 4 based on thresholding cosine similarity between vectors at 0.5.
\end{itemize}

\subsubsection{Use Analysis}
By clicking on the "Data" tab the user can (i) inspect uses or (ii) annotator judgments for each word in their projects (see Figure \ref{fig:concordances} in Appendix). Uses are displayed in concordance tables showing the aligned target words with additional context. The table offers sorting functions according to multiple criteria. Annotator judgments are displayed in a table with additional information such as the contexts of both uses, the data IDs, annotator name and annotator comment.

\subsubsection{Annotation statistics}
\label{sec:stats}
On the "Statistics" tab the user can calculate a range of summary statistics over the annotated data for analysis. These include 
\begin{inparaenum}[(i)]
   \setlength\itemsep{0pt}
   \item various measures of annotator agreement,
    \item label averages for words, groupings and annotators,
    \item comparisons between human and computational annotation.
\end{inparaenum}
Further, the system allows to infer various meta-measures from the basic semantic proximity  annotations. These include
\begin{inparaenum}[(i)]
   \setlength\itemsep{0pt}
    \item word sense clustering,
    \item semantic variation measures,
    \item semantic change measures.
\end{inparaenum}

\subsubsection{Annotation visualization}
\label{sec:visualization}

The visualization takes the form of a clustered network graph, calculated with Python in the backend, which the Pyvis visualization library, an interface for Vis.js, connects to a HTML, JavaScript and CSS frontend.\footnote{\url{https://www.ims.uni-stuttgart.de/data/wugs}} Nodes in the graph represent word uses and edge weights represent the annotations of use pairs (see Figure \ref{fig:visualization} in Appendix). Each edge's proximity score is calculated by taking the median between annotator judgments. When all of the annotators have given a score of "Cannot decide", the edge is marked with "NaN". If a node has at least half of its judgments as "Cannot decide", it is excluded from the clustering process.
The plotted graph can be filtered according to annotation group, date, edge weight, annotator and noise nodes or edges. The visualization also includes additional information, such as sense frequency and probability distributions, metainformation on uses, the clustering method and the node positioning method. Users can explore detailed information in additional "Stats" dropdowns. The main goal of the graph visualization is to make it easier to find information on word uses. In contrast to a large text document or table with hundreds of uses, in a clustered graph different meanings can be found quickly. 

\subsection{Frontend}
The user interface of DURel is designed with HTML, CSS, JavaScript and the Thymeleaf template engine. Thymeleaf is a natural choice for an application with Java Spring Boot as it provides full Spring Framework integration.

\subsection{Backend}
The DURel backend is built with the Spring framework and a PostgreSQL database, both of which are open-source and widely used in industry. 
It is responsible for user and project management, transferring project data (upload, download), handling the annotation process, and data analysis (use analysis, statistics). The backend runs the WUG visualization pipeline as a subprocess (see Section \ref{sec:visualization}).

The computational annotation pipeline is implemented as a separate component, built in Python with PyTorch\footnote{\url{https://pytorch.org/}} and Hugging Face\footnote{\url{https://huggingface.co}}: It retrieves annotation tasks users create on the DURel website, automatically generates annotations and sends them back to the DURel backend. We deploy the two components on different servers and let them interact with each other by sending REST API requests. The separation gives us the possibility to use the computational annotation pipeline independently from DURel with other annotation tools. It also allows us to deploy the computational annotation pipeline on any server, depending on computational workload, and to run multiple instances of the pipeline to spread the workload to multiple servers.

\section{Case Studies and Evaluation}
In this section, we describe two case studies to show the usefulness of the tool and to evaluate the computational annotators: (i) The \textit{arm} example is a small scale study on a chosen test word for which we selected word uses from a corpus. (ii) The lexicographer case study conducted on a set of 18 Swedish words is related to an ongoing lexicographic project. 

\subsection{The \textit{arm} example}
For various senses of the word \textit{arm}, \citet[][p. 41]{Schlechtweg2023measurement} selected the sentences (i.e. word uses) displayed in Table \ref{tab:corpus} from the online interface of COHA corpus \citep[][]{davies2002corpus}. We uploaded these uses to the DURel tool, which combined each sentence with every other sentence into use pairs. These pairs were then annotated with the XL-Lexeme annotator (see Section \ref{sec:pair}) and clustered using the correlation clustering algorithm. The resulting cluster structures were then visualized via the system's visualization function (see Figure \ref{fig:armtool} in Appendix). The generated cluster structures are very similar to the manual annotations: both clusterings distinguish the metaphorical sense `arm of the sea' (recall Section \ref{sec:background}). However, the computational annotator merges the `body part' and the `weapon' sense. Hence, while the computational annotator is not perfect, it works reasonably well for this example. We hope that this annotator will prove useful to discover meaning structures in various meaning-related study areas. One such application is given in the next subsection. (We also provide a video demonstration of this example.\footnote{\url{https://www2.ims.uni-stuttgart.de/video/durel-tool/230623-durel-tool-demo.mp4}.})

\subsection{The lexicographer case study}
To test DURel in a practical application, we used it in a small study on revealing semantic variation in Swedish. The study is closely related to the work of revising the comprehensive Swedish dictionary published by the Swedish Academy ('The Contemporary Dictionary of the Swedish Academy'; in short SO\footnote{\url{https://svenska.se/}}), a general language definition dictionary with about 65,000 headwords. 
An important part of the revision work within the dictionary project is to examine whether the meanings of the headwords in SO have developed in some way since the last edition was published. 
However, the lexicographic team of SO currently do not, in a systematic way, use any formal computational methods for discovering semantic change on the lexical level. Hence, the lexicographic challenge is finding relatively new meanings not already recorded. The question is whether the DURel tool can point out meaning variation (as an indicator of change) by clustering different meanings of a word. For that purpose, as a first round of experiments, we selected a set of 18 established Swedish words that are already in SO. These words were thoughtfully selected based on their semantic characteristics: all of them have a main sense and one or more subsenses. For each word, we automatically extracted a collection of 50 random sentences from the SVT (Swedish Television) corpus available through Korp \cite{Borin-Lars2012-156080}. These sentences were then uploaded into the tool, automatically paired and annotated with XL-Lexeme, and finally clustered using the correlation algorithm. The resulting clusters were evaluated against lexicographer judgments for sense clusters (gold data). 

\begin{table}[t]
\small
\centering
\begin{tabular}{l|l|l|l}
\toprule
\textbf{Word} & \textbf{ARI} & \textbf{Word} & \textbf{ARI} \\ \midrule
ofantlig & 1.0  & klimat & 0.083   \\ 
enkelspårig & 1.0 & vansinnig & 0.0  \\ 
baksida & 0.912 & lirka & 0.0  \\ 
bagage & 0.785 & kapitulera & 0.0 \\ 
fasad & 0.652 & hemmaplan & 0.0   \\ 
vissen & 0.645 & hagla & 0.0  \\ 
skör & 0.507 & fotavtryck & 0.0  \\ 
rutten  & 0.333 & tvärnita & -0.019  \\ 
ventilera & 0.303 & kriga & -0.025   \\
\hline
 \textbf{Average}& \multicolumn{3}{c}{\textbf{0.343}}\\
 \bottomrule
\end{tabular}
\caption{Cluster evaluation based on ARI.}
\label{tab:eval-results}
\end{table}

Table \ref{tab:eval-results} displays the evaluation results using the Adjusted Rand Index \citep[ARI,][]{Hubert/Arabie:85} to compare automatically generated clusters with manually curated gold clusters. Each row represents a specific word used to form the clusters, with the corresponding ARI value indicating the similarity between the automatic and gold clusters generated by DURel and the lexicographer respectively. ARI generally ranges from -0.5 to 1, where higher values closer to 1 signify better agreement between the gold clusters and the derived clusters. A value of 0 suggests a random clustering. As can be noted words like \textit{ofantlig}, \textit{enkelspårig}, \textit{baksida}, and \textit{bagage} demonstrate relatively high ARI values, indicating stronger alignment between their respective clusters and the gold clusters. In contrast \textit{tvärnita}, \textit{hemmaplan} and \textit{vansinnig} show ARI values of 0 or slightly negative. The `Average' row provides an overall assessment by presenting the average ARI across all words. The moderately positive average value suggests that the automatically derived clusters generally encode meaningful semantic information which could be useful for lexicographic work.

\begin{figure}[t]
    \centering
    \includegraphics[scale=0.3]{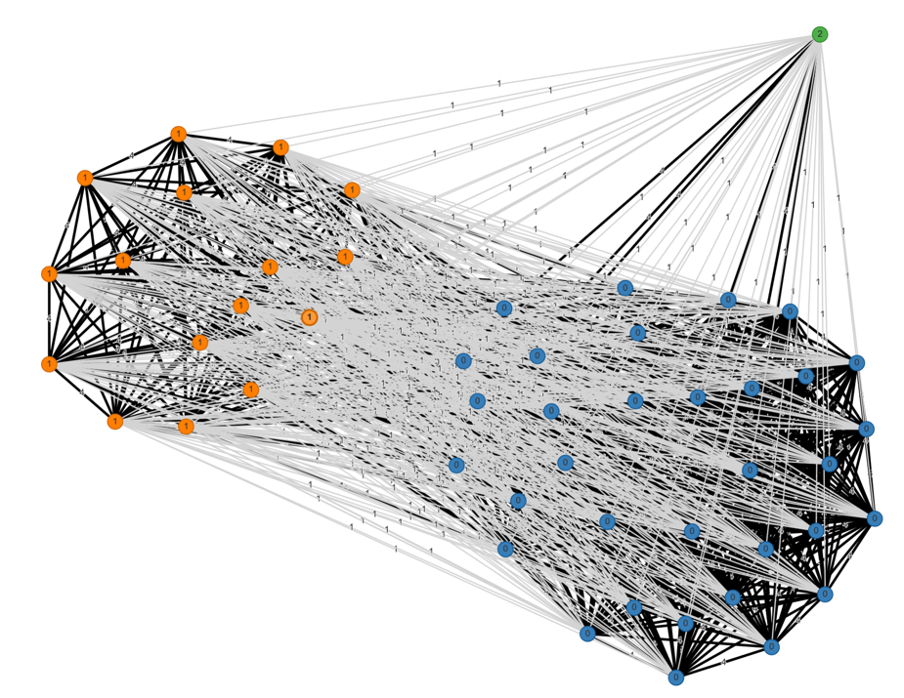}
    \caption{Clusters for \textit{bagage}.}
    \label{fig:baggage}
\end{figure}

In addition to automatic evaluation using ARI, the clusters were analyzed qualitatively by a lexicographer, which provided useful insights. For example consider Figure \ref{fig:baggage} which shows the clusters produced for the word \textit{bagage} (Eng: `luggage'). As can be seen, the uses were clustered into three main clusters (colored blue, orange, and green). The blue and orange colored represent the literal and figurative usage of the word respectively while one of the uses was wrongly put into the third, green cluster. A discrepancy between the number of clusters identified by DURel and the number of senses recorded in SO dictionary for the candidate word can be used as an indicator that the description in the dictionary is outdated.
Such words can be prioritized for manual inspection in order to update the dictionary entry or otherwise to improve the computational model. An examination of the figurative examples (the orange cluster) revealed that some of the uses can be classified as different variants of the same Swedish idiomatic expression (e.g. \textit{ha något i bagaget}, ENG: `to have something in the luggage'). This particular idiom, and similar others, is not treated in SO 2021 and will be included in the next edition. Further, the text examples from the corpora form also a good basis to include more language examples in the SO dictionary. In summary, the DURel tool was evaluated by the lexicographer to be helpful in revising the SO dictionary, who otherwise has relied completely on manual cumbersome methodology.  

\section{Conclusion}
\label{sec:conclusion}
We presented the DURel, an online, open-source annotation tool for annotating semantic proximity between word uses. The tool supports 
human as well as computational annotation, building on recent advances with WiC models. Annotator judgments are automatically clustered and visualized for analysis. This allows to capture word senses with simple and intuitive micro-task judgments between use pairs, requiring minimal preparation efforts. Additionally, DURel can compare the agreement between annotators to guarantee the inter-subjectivity of the obtained judgments and to calculate summary statistics over the annotated data giving insights into sense frequency distributions, semantic variation or changes of senses over time. The computational annotator component allows to generate word sense clusters for large sets of words and word uses, making it possible, for instance, to analyze large amounts of data or to search systematically for new senses.

A number of different research groups around the world have used the system to annotate data \citep[e.g.][]{Zamora2022lscd,kutuzov2022nordiachange,Aksenova2022rudsi,Chen2023chiwug} and continue to do so: Currently, the tool is being used to annotate an Italian dataset.

\subsection{Limitations}
DURel provides a range of functionalities feasible only because it focuses on a single task but with the disadvantage of a more narrow application range and low customization. This also binds users to the four-level annotation scale hard-coded into the system. While this scale is motivated by theory \citep[][]{Schlechtwegetal18}, there exist valid alternatives \citep[e.g.][]{Erk13,Brown2008}. For users who want to use such an alternative scale, we recommend to conduct the annotation study within a more customizable annotation system like PhiTag. The annotated data can then be uploaded to DURel with the "Upload Judgments" functionality, and clustering and analysis can be applied within DURel.

The tool currently allows to specify a continuous substring of any length as target string in each uploaded word use. Discontinuous target strings, as needed e.g. for discontinuous multi-word expressions such as particle verbs in German, are currently not allowed.

The computational annotators are currently restricted to predict binary labels. In the future, we will provide WiC models optimized for ordinal predictions on all levels of the annotation scale \citep[][]{Zhang2023thesis}.

\section*{Acknowledgments}
The development of the tool has been funded by the project `Towards Computational Lexical Semantic Change Detection' supported by the Swedish Research Council (2019–2022; contract 2018-01184), the research program `Change is Key!' supported by Riksbankens Jubileumsfond (under reference number M21-0021) and the CRETA center funded by the German Ministry for Education and Research (BMBF). We thank Anne Reuter, Annalena Streichert, Enrique Medina Waldo Castaneda, Sinan Kurtyigit, Serge Kotchourko, Pedro G. Bascoy, Francesco Periti and Jing Chen for contributions to this project.

\bibliography{Bibliography-general,bibliography-self,bibliography-supervision-self,additional-references}
\bibliographystyle{acl_natbib}

\newpage
\appendix

\section{Appendix}
\label{sec:appendix}
We have included some screenshots of the DURel application below.

\begin{figure*}
    \centering
    \includegraphics[width=\linewidth]{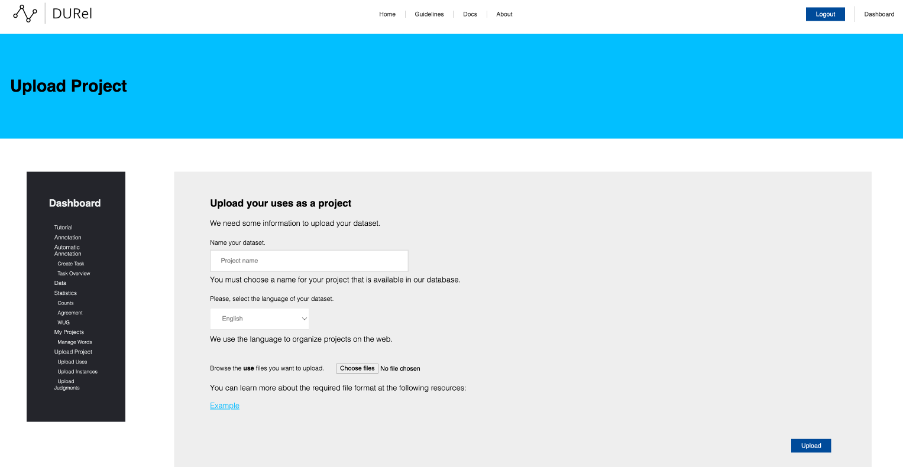}
    \caption{Upload uses tab: Upload interface for word uses.}
    \label{fig:upload}
\end{figure*}

\begin{figure*}
    \centering
    \includegraphics[width=\linewidth]{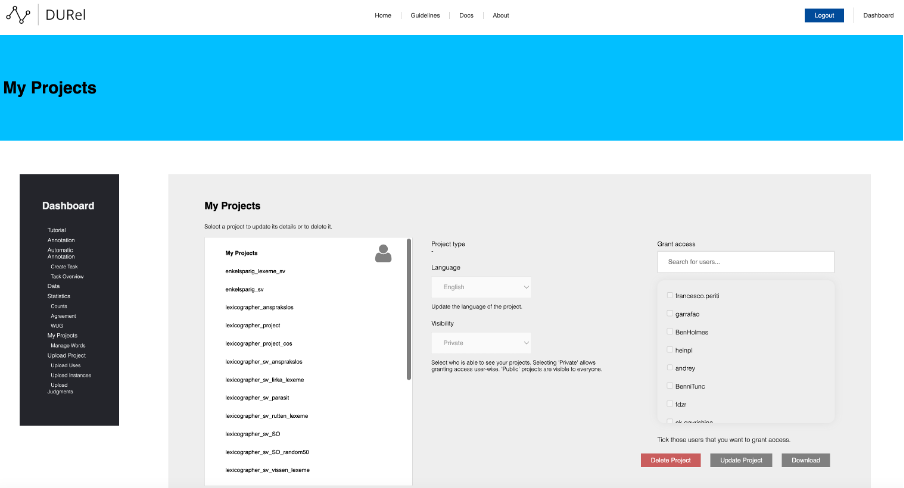}
    \caption{My Projects tab: Assign access rights to annotators, delete or download projects.}
    \label{fig:projects}
\end{figure*}

\begin{figure*}[p]
\centering
    \centering
    \includegraphics[width=\linewidth]{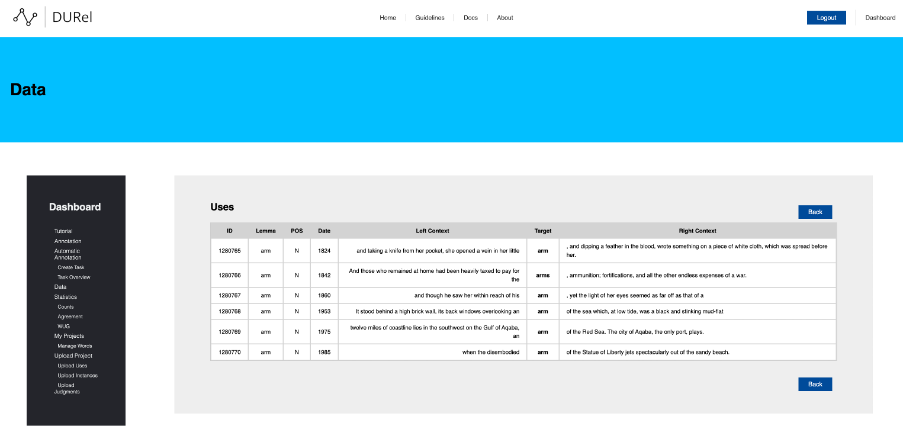}
    \caption{Data tab: Shows concordances in a table.}
    \label{fig:concordances}
\end{figure*}

\begin{figure*}[p]
\centering
    \centering
    \includegraphics[width=\linewidth]{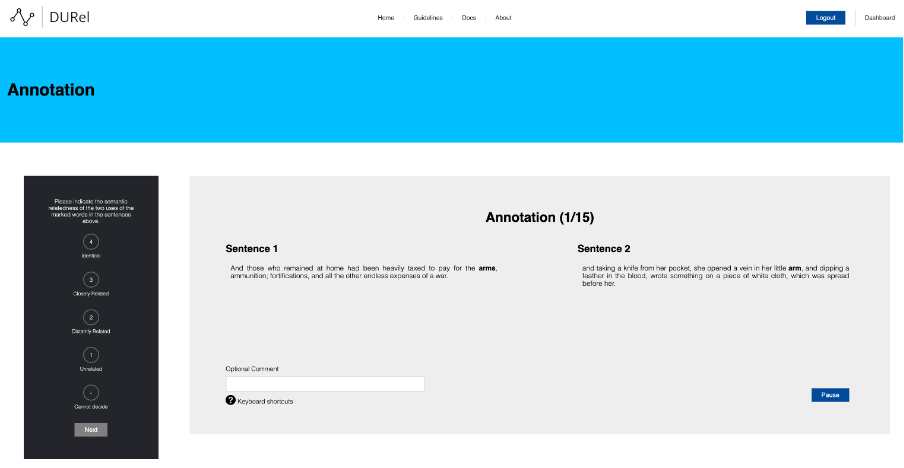}
    \caption{Annotation interface: Annotation instance (use pair) presented to annotator.}
    \label{fig:annotate}
\end{figure*}

\begin{figure*}
    \centering
    \includegraphics[width=\linewidth]{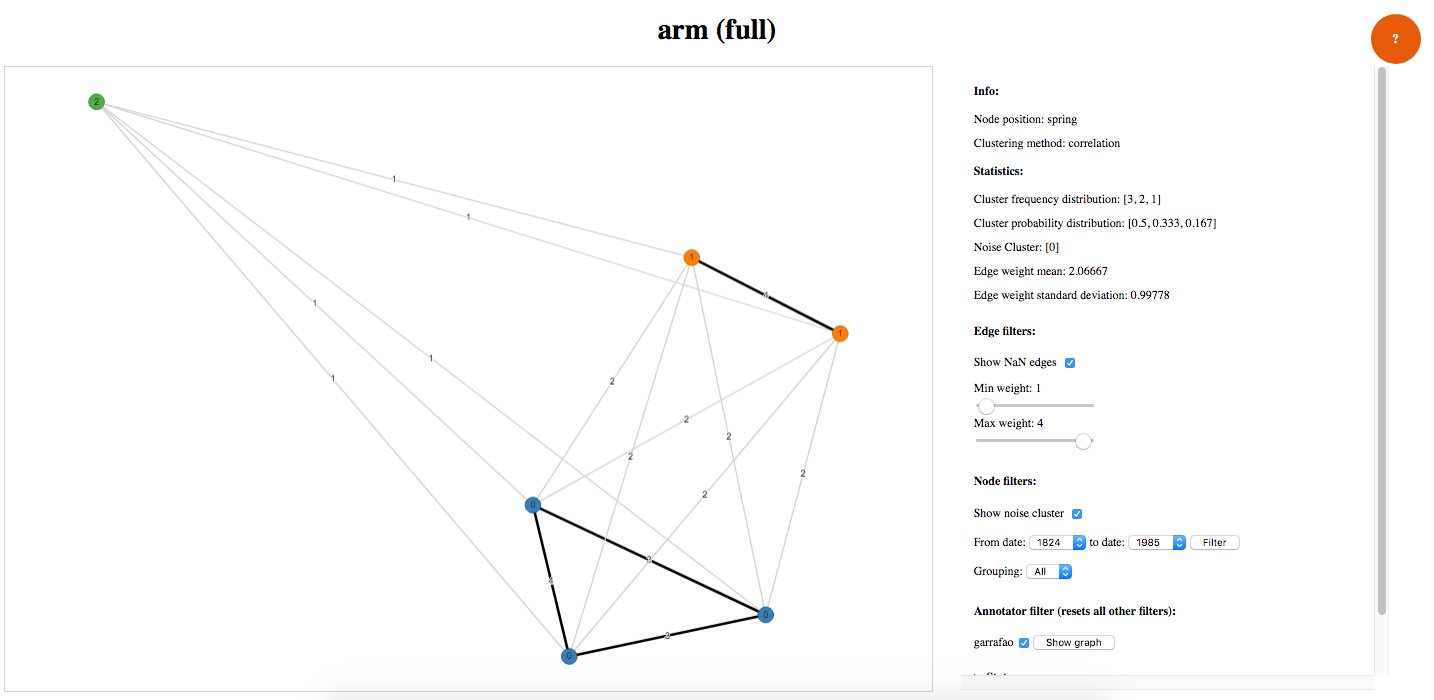}
    \caption{Visualization: Result of visualization pipeline available on "Statistics" tab.}
    \label{fig:visualization}
\end{figure*}

\begin{figure*}[t]
   \begin{subfigure}{0.5\textwidth}
\frame {        \includegraphics[width=\linewidth]{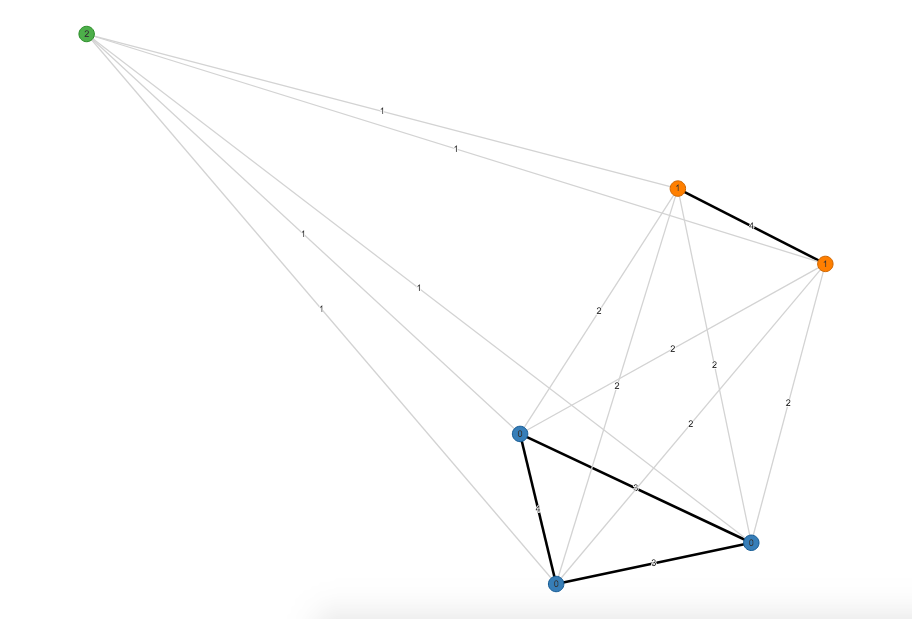}}
        \caption*{human}
      \end{subfigure}
    \begin{subfigure}{0.5\textwidth}
\frame{        \includegraphics[width=\linewidth]{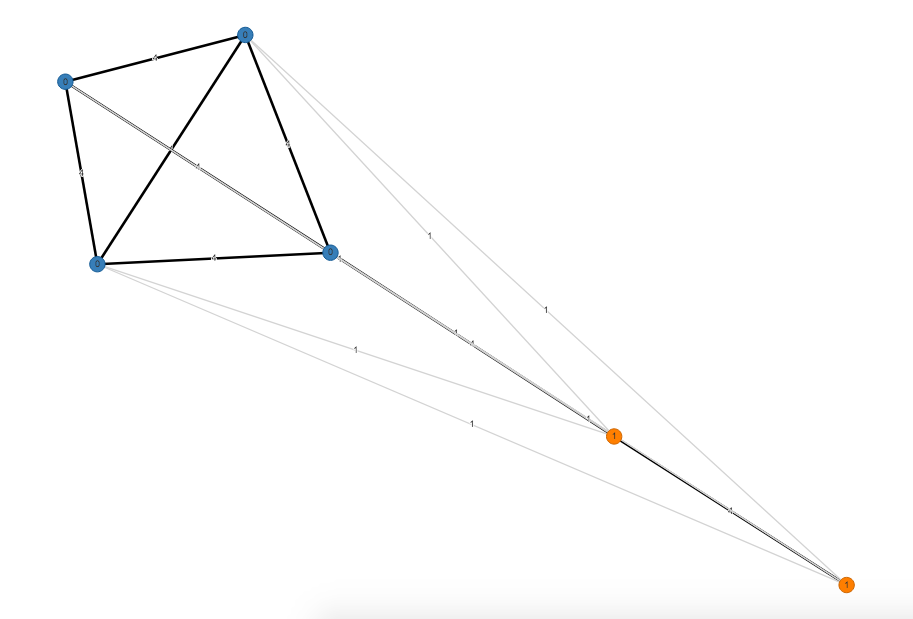}}
        \caption*{computer}
    \end{subfigure}
   \caption{Clustering obtained from human vs. computational annotations of \textit{arm} in DURel tool. Orange cluster corresponds to metaphorical sense `arm of the sea' in both cases. Computational annotator merges `body part' and `weapon' sense.}
   \label{fig:armtool}
\end{figure*}

\end{document}